\documentclass{article}
\usepackage{spconf,amsmath,graphicx,hyperref}

\usepackage{algorithm}
\usepackage{algorithmic}
\usepackage{booktabs}
\usepackage{multirow}
\usepackage{makecell}
\usepackage{threeparttable}
\usepackage{amsmath}
\usepackage{amssymb}
\usepackage{enumitem} 
\usepackage{tabularx}

\setlist[itemize]{
    itemsep=0pt,        
    parsep=0pt,         
    leftmargin=1.33em,
    topsep=0pt,
}

\usepackage{caption}
\title{Artifact-Aware Evaluation for High-Quality Video Generation}
\name{
\begin{tabular}{c}
Chen Zhu$^{1,2}$\textsuperscript{*}\thanks{\textsuperscript{*}Equal Contribution}, Jiashu Zhu$^{2}$\textsuperscript{*}, Yanxun Li$^{2}$, Meiqi Wu$^{2,3}$, Bingze Song$^2$, \\
Chubin Chen$^{2}$, Jiahong Wu$^{2}$\textsuperscript{\dag}\thanks{\textsuperscript{\dag}Corresponding authors}, Xiangxiang Chu$^{2}$, Yangang Wang$^{1}$\textsuperscript{\dag}
\end{tabular}
}

\address{
$^{1}$ Southeast University \ $^{2}$ AMAP, Alibaba Group \ $^{3}$ CASIA}

\begin{document}

\maketitle

\begin{abstract}
With the rapid advancement of video generation techniques, evaluating and auditing generated videos has become increasingly crucial. Existing approaches typically offer coarse video quality scores, lacking detailed localization and categorization of specific artifacts. In this work, we introduce a comprehensive evaluation protocol focusing on three key aspects affecting human perception: Appearance, Motion, and Camera. We define these axes through a taxonomy of 10 prevalent artifact categories reflecting common generative failures observed in video generation. To enable robust artifact detection and categorization, we introduce GenVID, a large-scale dataset of 80k videos generated by various state-of-the-art video generation models, each carefully annotated for the defined artifact categories. Leveraging GenVID, we develop DVAR, a Dense Video Artifact Recognition framework for fine-grained identification and classification of generative artifacts. Extensive experiments show that our approach significantly improves artifact detection accuracy and enables effective filtering of low-quality content.
\end{abstract}


\section{Introduction}
\label{sec:Introduction}

Recent advances in video generation have enabled broad deployment across creative and industrial domains, leading to a rapid proliferation of synthetic content. Despite impressive progress~\cite{chen2025s2guidancestochasticselfguidance, wu2025imagerysearch}, even state-of-the-art (SOTA) models~\cite{wan2.1, ma2025step} often exhibit subtle yet consequential artifacts—flickering textures, temporal discontinuities, non-physical dynamics, and unstable camera motion. These instabilities, especially in user-produced AI content, incur high manual moderation costs on platforms. Thus, developing scalable and efficient methods to evaluate and audit generated videos is an essential and timely research problem.

Traditional video quality assessment (VQA) metrics such as SSIM and VMAF rely on hand-crafted, signal-statistics-based features and perform well under controlled, full-reference settings but generalize poorly to the heterogeneous distortions in user-generated content (UGC)~\cite{zheng2024video}. To overcome this limitation, recent works~\cite{yuan2024ptm, mitra2025vision} leverage vision–language models (VLMs) like CLIP to move beyond pixel-level cues toward semantic understanding. Building on this, the current frontier of VQA incorporates multimodal large language models (MLLMs)~\cite{wen2025ensemble, he2024videoscore}, enabling end-to-end quality assessment through instruction tuning or prompting. However, most MLLM-based approaches~\cite{wang2025unified, liu2025improving} regress coarse-grained scores, which lacks interpretability and fails to account for fine-grained quality factors.

Although modern generative models can produce strikingly realistic videos, they inevitably introduce artifacts. Human viewers primarily evaluate video quality based on the presence and severity of such artifacts—for example, a video with large-scale warping is immediately dismissed—rather than abstract, aggregate quality scores. Consequently, video evaluation should move away from monolithic scores toward structured, fine-grained assessments aligned with human perception. To address this, we propose a perceptually grounded evaluation protocol, structured along three axes—Appearance, Motion, and Camera—that capture artifacts commonly regarded as objectionable. Within this framework, we define 10 recurrent artifact categories observed across video generation systems, encompassing spatial appearance (e.g., texture corruption, object deformation), temporal dynamics (e.g., flicker, motion discontinuities), and camera behavior (e.g., unstable trajectories, implausible parallax). This structured methodology enables precise artifact identification and facilitates human-aligned quality assessments.

\begin{figure*}[htbp]
    \centering
    \includegraphics[width=\textwidth]{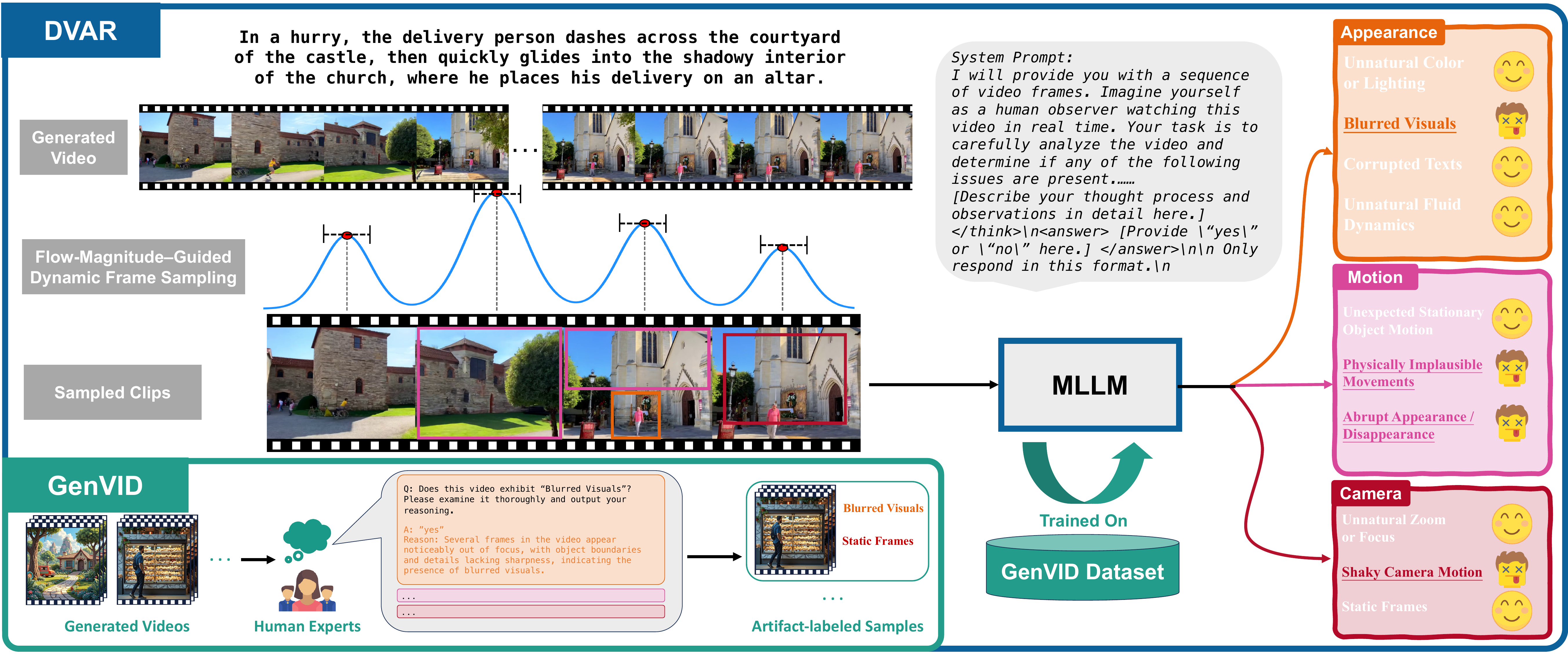}
    \caption{Overview of the artifact-aware evaluation pipeline, including GenVID data annotation and the inference workflow of DVAR. Given an input video, FMG-DFS is first employed to sample video clips with a high probability of containing artifacts. These clips are then subjected to inference by a trained MLLM to identify the presence of each potential artifact category.}
    \label{fig:pipeline}
    \vspace{10pt}
\end{figure*}

However, current MLLMs lack the capacity to reliably analyze video artifacts, as their pretraining largely relies on natural, artifact-free imagery~\cite{li2024llava, bai2025qwen2, li2023videochat}. To bridge this gap, we curate \textbf{GenVID}, a dataset of 80k videos produced by popular generative models and meticulously annotated across the 10 artifact categories. GenVID provides the diversity and scale necessary for training and benchmarking artifact detection models. Moreover, existing models often employ uniformly sampled frames, which fail to capture transient defects caused by temporal dynamics. To address this, we introduce Flow-Magnitude-Guided Dynamic Frame Sampling (\textbf{FMG-DFS}), which focuses on motion-salient intervals to improve localization and detection efficiency. Building on GenVID, we propose \textbf{DVAR}, a Dense Video Artifact Recognition framework that couples FMG-DFS with fine-grained artifact detection. DVAR bridges the gap between coarse score-based assessments and explicit artifact recognition, enabling systematic filtering of low-quality content and providing actionable feedback for moderation workflows.

Extensive experiments demonstrate that our approach significantly improves artifact recognition over baselines. When deployed as a filtering module, DVAR effectively suppresses low-quality video outputs while preserving high-quality, diverse generations, making it a scalable solution for real-world pipelines. Our contributions can be summarized as follows:

\begin{itemize}
    \item A human-centric evaluation protocol that structures video quality along three perceptual axes—Appearance, Motion, and Camera, and operationalizes 10 prevalent artifact categories.
    \item GenVID, a large-scale dataset of 80k generated videos from popular generation model, meticulously annotated at the category level for all artifacts, supporting robust training and fair benchmarking.
    \item DVAR, a dense video artifact recognition framework that couples FMG-DFS strategy with fine-grained detection, enabling precise and scalable filtering.
\end{itemize}

\section{Proposed Method}
\label{sec:Proposed Method}
This section outlines our approach. We begin by formalizing an artifact-centric evaluation protocol structured along three perceptual axes and comprising 10 artifact categories~(Sec.~\ref{subsubsec:Artifact-Centric Evaluation Protocol}). Next, we leverage GenVID for annotation and benchmarking~(Sec.~\ref{subsubsec:Dataset Construction}). To efficiently capture transient defects, we introduce a FMG-DFS strategy that focuses on motion-salient intervals~(Sec.~\ref{subsubsec:Flow-Magnitude–Guided dynamic frame sampling}). Finally, we detail the DVAR framework for fine-grained artifact detection with robust performance~(Sec.~\ref{subsubsec:Dense Video Artifact Recognition}).

\subsection{Artifact-Centric Evaluation Protocol}
\label{subsubsec:Artifact-Centric Evaluation Protocol}
Generated videos inevitably contain a variety of artifacts, and both the types of artifacts and individual sensitivity to them shape the viewing experience. To ground evaluation in human perception rather than abstract indices, we aim to define a fair, consensus-oriented taxonomy of artifacts that reflects aggregate viewer preferences. To this end, we first assembled a seed datasets by generating videos from 10k diverse prompts spanning multiple domains using state-of-the-art video generation models. We then recruited 20 raters with diverse age, gender, and occupational backgrounds to evaluate the clips. Raters assessed overall viewing quality and, whenever a clip was deemed unsatisfactory, were required to provide a concrete reason. 

\begin{figure}[htbp]
    \centering
    \includegraphics[width=\columnwidth]{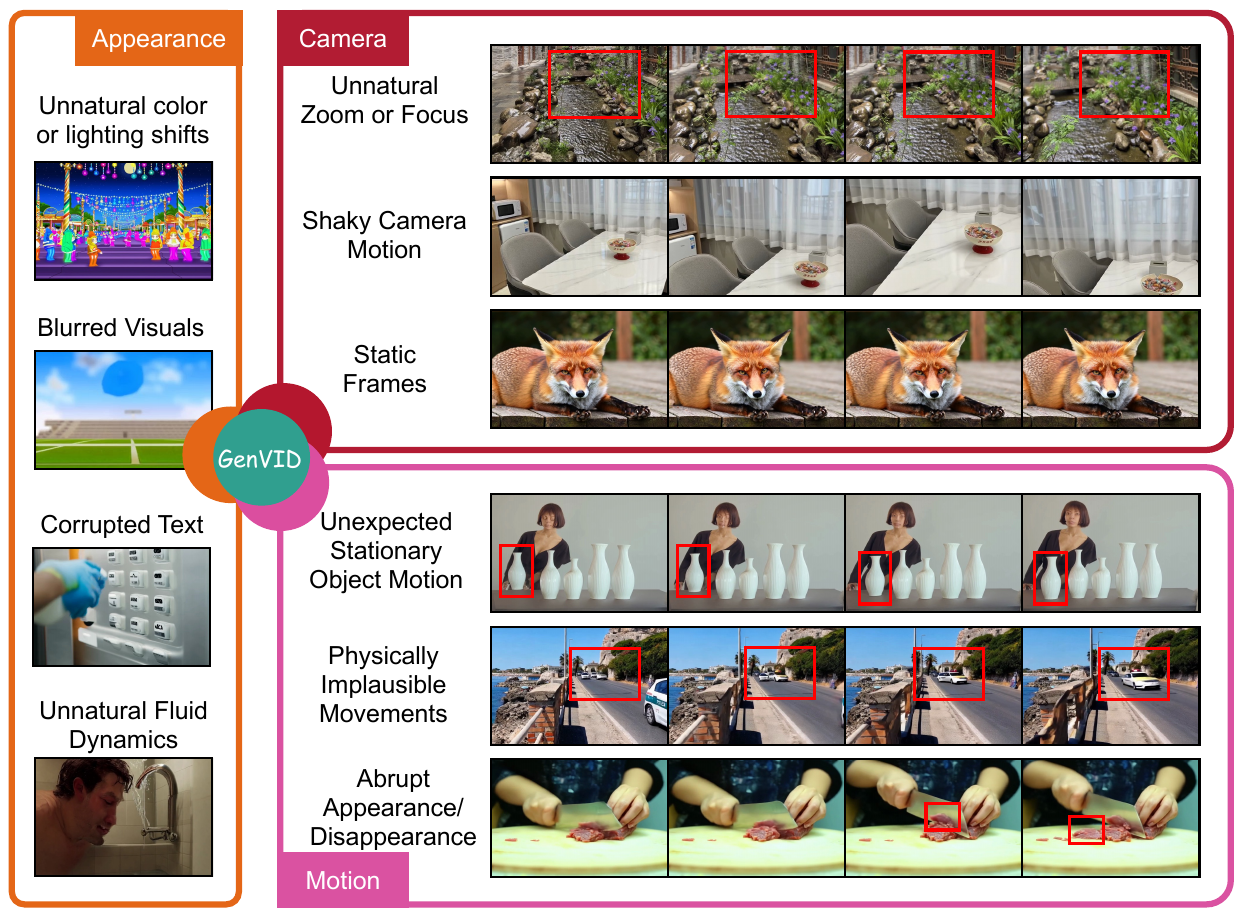}
    \caption{Examples of GenVID dataset. Each generated video containing noticeable artifacts is subjected to human annotation and categorization.}
    \label{fig:cases}
    \vspace{-10pt}
\end{figure}

Analyzing these judgments and rationales, we distilled three perceptual axes along which generative failures most commonly manifest: Appearance, Motion, and Camera. Within each axis, we selected the four most salient artifact categories based on prevalence across models and prompts and on perceived severity to viewers, yielding a 10-category taxonomy~(shown in Fig.~\ref{fig:cases}). This taxonomy serves as the foundation of our evaluation protocol and aligns quality assessment with the issues users most frequently find objectionable.

\subsection{Construction of GenVID}
\label{subsubsec:Dataset Construction}
\noindent\textbf{Videos Generation.} To ensure diversity, we generate 100k clips using multiple popular open-source video generation models—Wan 2.1~\cite{wan2.1}, CogVideoX~\cite{yang2024cogvideox}, and Open-Sora\cite{zheng2024opensora}, covering a broad range of scenes, prompts, resolutions, and frame counts. We perform an initial cleaning pass to discard invalid outputs (e.g., blank, garbled, or trivially failed generations). After filtering, we retain 80k valid videos that span heterogeneous content.

\noindent\textbf{Annotation and QA formulation.} Each retained video is evaluated against the 10 artifact categories defined in our protocol. As shown in the bottom panel of Fig.~\ref{fig:pipeline}, raters determine the presence or absence of each artifact category for every video using a guideline document containing definitions and examples. Inspired by VisionReward~\cite{xu2024visionreward}, we cast these multi-label annotations into a Question–Answer~(QA) format by pairing each video with category-specific prompts (e.g., "$Does \ this \ video\ exhibit \ \{Artifact\} \ ?$") and binary labels. With lightweight augmentations, we yield 960k QA pairs over the 80k videos. This design both preserves the granularity of artifact labels and facilitates training and evaluation of models under a unified, scalable supervision format.

\subsection{Flow-Magnitude–Guided Dynamic Frame Sampling}
\label{subsubsec:Flow-Magnitude–Guided dynamic frame sampling}

Videos often contain redundant information. Current MLLMs typically process videos by dividing them into sequential image frames, which are encoded by a visual encoder and fused with textual inputs before model inference. However, densely frames sampling~\cite{he2024videoscore} significantly increases computational costs, making it impractical to handle videos with longer durations or higher resolutions under constrained GPU memory. On the other hand, sparsely sampling frames risks omitting critical information. To address this, AKS~\cite{tang2025adaptive} proposed a method that leverages multimodal features to calculate the relevance scores between video frames and textual inputs, enabling importance-based frame sampling. However, this approach requires the text to specify specific content within the video frames, which renders it ineffective for open-ended queries (e.g., "What does this video depict?") where relevance scores are less informative.

\begin{algorithm}[ht]
\caption{\ FMG-DFS}
\label{algo:FMG-DFS}
\begin{algorithmic}[]
\REQUIRE Video frames $V = \{f_1, ..., f_n\}$; segments $K$; peak distance $W$; sample count $M$; smoothing window $w_s$
\ENSURE Sampled indices $C$

\STATE Compute instability scores $S$ for $V$ using optical flow
\STATE $S_{smooth} \leftarrow \text{Smooth}(S, w_s)$
\STATE $P \leftarrow \text{FindPeaks}(S_{smooth}, W)$

\STATE $P_{top} \leftarrow \text{TopK}(P, K)$; \, $F_{per} \leftarrow \lfloor M/K \rfloor$; \, $C \leftarrow \emptyset$

\FOR{peak $p$ in $P_{top}$}
    \STATE $start \leftarrow \max(0, p-F_{per}/2)$;
    \STATE $end \leftarrow \min(n, start + F_{per})$
    \STATE $C.\text{append}((start, end))$
\ENDFOR

\STATE $C \leftarrow \text{AdjustClips}(C, M)$; \, $C \leftarrow \text{RemoveOverlap}(C)$
\IF{$\text{TotalFrames}(C) < M$}
    \STATE $C \leftarrow \text{FillGaps}(C, M)$
\ENDIF

\RETURN $C$
\end{algorithmic}
\end{algorithm}

To overcome these challenges, we propose FMG-DFS strategy. Our key insight is that video segments with pronounced motion dynamics are more likely to exhibit artifacts and temporal inconsistencies. 
Accordingly, we compute dense optical flow~\cite{farneback2003two} for each frame and use flow magnitudes to sample frames at peaks of motion intensity (see Algorithm~\ref{algo:FMG-DFS}). This lightweight approach balances computational efficiency and information density. By targeting motion-salient segments, FMG-DFS enables efficient inference and precise localization of temporal regions most pertinent to video-related queries.

\subsection{Dense Video Artifact Recognition}
\label{subsubsec:Dense Video Artifact Recognition}

To address the challenges of understanding and identifying artifacts in generated videos, we propose a framework called Dense Video Artifact Recognition (see Fig.~\ref{fig:pipeline}). Training MLLM with GenVID enables them to develop a comprehension of how video artifacts manifest. Meanwhile, the proposed FMG-DFS enhances the efficiency of video frame sampling, aiding the temporal localization of artifacts with higher precision. By combining these two components, DVAR is designed to detect and classify artifacts in generated videos in a systematic manner. This synergy not only facilitates efficient artifact detection but also enables accurate categorization of different types of video artifacts through targeted sampling and robust model training.

\section{Experiments}
\label{sec:Experiments}

\subsection{Experimental Setup}
\label{subsec:Experimental Setup}

\noindent\textbf{Implementation Details.}
All experiments were conducted on a machine equipped with 8 NVIDIA H20 GPUs. We selected Qwen2.5-VL-7B~\cite{bai2025qwen2} as the baseline model. During training, the visual encoder and the multimodal projection layers were kept frozen, while only the language model component was fine-tuned. 

\noindent\textbf{Metrics.} Model artifact recognition is measured by accuracy (ACC), defined as $\text{ACC} = \frac{1}{N} \sum_{i=1}^{N} \mathbb{I}(\text{pred}_i = \text{gt}_i)$, where $\text{pred}_i$ and $\text{gt}_i$ are the predicted and ground-truth labels for the $i$-th video, respectively. The indicator function $\mathbb{I}(\cdot)$ equals 1 if the prediction matches the ground truth, and 0 otherwise. All dimensional metrics (\textit{Appearance}, \textit{Camera}, \textit{Motion} and \textit{All}) in the tables are computed using ACC.

\subsection{Comparison with SOTA Models}
\label{subsec:Comparison with SOTA Models}
In this section, we evaluate the performance of DVAR in comparison with SOTA open-source and proprietary models on the task of video artifact detection. To ensure a fair comparison, 10 frames are uniformly sampled from each video as input, and a single query is posed for each sample. The prompts instruct the models to provide binary answers, specifically "yes" or "no" with the final response extracted using regular expressions. Outputs that are off-topic, irrelevant, or repetitive are considered incorrect responses and classified as failures.

\begin{table}[htbp]
\centering
\caption{Comparison of artifact recognition accuracy on the GenVID testset.}
\label{tab:comparasion}
\resizebox{\linewidth}{!}
{\begin{tabular}{@{}ll|cccc@{}}
\toprule
Method                           & Params & Appearance & Camera & Motion & All   \\ \midrule
GPT-5                            & -              & 0.668      & 0.625  & 0.721  & 0.672 \\
GPT-4o                           & -              & 0.481      & 0.406  & 0.526  & 0.471 \\
LLaVA-NeXT~\cite{li2024llava}    & 7B             & 0.489      & 0.475  & 0.455  & 0.473 \\
VideoChat~\cite{li2023videochat} & 7B             & 0.560      & 0.523  & 0.532  & 0.538 \\
Qwen2.5-VL~\cite{bai2025qwen2}   & 3B             & 0.462      & 0.449  & 0.505  & 0.472 \\
Qwen2.5-VL~\cite{bai2025qwen2}   & 72B            & 0.537      & 0.501  & 0.630  & 0.556 \\ \midrule
\textbf{DVAR-Mean-7B}                    & 7B             & \textbf{0.849}      & \textbf{0.785}  & \textbf{0.767}  & \textbf{0.800} \\ \bottomrule
\end{tabular}}
\end{table}

The results in Tab.~\ref{tab:comparasion} demonstrate that our model outperforms other MLLMs in the task of video artifact recognition. It is evident that both proprietary and open-source large-scale models lack sufficient understanding of video artifacts across various dimensions. By undergoing supervised fine-tuning on the GenVID dataset, the MLLM successfully learned to identify artifacts present across different dimensions in generated videos. Notably, in the \textbf{Appearance} dimension, DVAR achieved a significant improvement in accuracy compared to the best-performing model, showcasing the model’s ability to capture spatial features of pixel-level artifacts. Given that artifacts inherently manifest as disturbances, learning temporal characteristics poses greater challenges. Nevertheless, DVAR achieved approximately a 4\% improvement in the \textbf{Motion} dimension, further demonstrating its effectiveness.

\subsection{Ablation Studies}
\label{subsec:Ablation Studies}
To further investigate the contribution of each module to the performance improvements of DVAR, we designed a series of ablation studies focusing on the base model parameter size and video frame sampling strategies. For all input videos, a fixed 10-frame sampling strategy was applied: \textbf{Random} denotes random sampling, while \textbf{Mean} refers to uniform sampling. Regarding \textbf{AKS} method, the configuration used was as follows: $ratio=1$, $t_1=0.8$, $t_2=-100$, and $depth=3$. For \textbf{FMG-DFS} method, the parameters were set to $K=3$, $W=5$, and $M=10$. For each sampling strategy, fine-tuning was conducted on two base models: Qwen2.5-VL-3B and Qwen2.5-VL-7B, to evaluate their respective capabilities and performance improvements.

As shown in Tab.~\ref{tab:ablation}, our FMG-DFS strategy demonstrates superior efficiency in capturing potential artifact frames compared to uniform sampling, thereby further enhancing the performance of DVAR. Moreover, increasing the model parameter size provides only limited improvements (approximately 1\%) in learning artifact features, underscoring the substantial gains already achieved through our training on the GenVID dataset.

\begin{table}[]
\caption{Ablation study of sampling strategies and model sizes for artifact detection.}
\label{tab:ablation}
\resizebox{\linewidth}{!}
{\begin{tabular}{@{}ll|cccc@{}}
\toprule
Sampling Strategy
& Params & Appearance & Camera & Motion & All   \\ \midrule
\multirow{2}{*}{Random}                      & 3B             & 0.722      & 0.695  & 0.721  & 0.713 \\
                                             & 7B             & 0.752      & 0.701  & 0.692  & 0.714 \\
\multirow{2}{*}{Mean}                        & 3B             & 0.841      & 0.763  & 0.749  & 0.784 \\
                                             & 7B             & 0.849      & 0.785  & 0.767  & 0.800 \\
\multirow{2}{*}{AKS~\cite{tang2025adaptive}} & 3B             & 0.816      & 0.797  & 0.732  & 0.782 \\
                                             & 7B             & 0.832      & 0.805  & 0.745  & 0.794 \\ \midrule
\multirow{2}{*}{\textbf{FMG-DFS}}                     & 3B             & 0.874      & 0.815  & 0.793  & 0.827 \\
                                             & 7B             & \textbf{0.876}      & \textbf{0.838}  & \textbf{0.819}  & \textbf{0.844} \\ \bottomrule
\end{tabular}}
\end{table}

\section{Conclusion}
\label{sec:Conclusion}

In this paper, we proposed a comprehensive artifact-aware evaluation framework for video generation, focusing on fine-grained artifact detection across Appearance, Motion, and Camera dimensions. We introduced the large-scale GenVID dataset and the novel DVAR framework, advancing the state-of-the-art in dense video artifact recognition. Experimental results demonstrate the accuracy of our approach in identifying artifacts in generated videos. Additionally, the FMG-DFS strategy enhances temporal localization, enabling more precise and efficient artifact detection. Extensive experiments demonstrate that our method accurately identifies generative artifacts, facilitating the efficient filtering of low-quality videos.

\clearpage
\bibliographystyle{IEEEbib}
\bibliography{main}

\end{document}